\documentclass{article}
\usepackage{PRIMEarxiv}

\usepackage[utf8]{inputenc} 
\usepackage[T1]{fontenc}    
\usepackage{chessfss}
\usepackage{hyperref}       
\usepackage{url}            
\usepackage{booktabs}       
\usepackage{amsfonts}       
\usepackage{nicefrac}       
\usepackage{microtype}      
\usepackage{lipsum}
\usepackage{graphicx}       
\usepackage{wrapfig}
\usepackage{xcolor}
\graphicspath{{media/}}     
\usepackage{tablefootnote}
\usepackage{pifont}
\usepackage{multirow}
\usepackage{textcomp}
\usepackage{amsmath}
\usepackage{amssymb}
\usepackage{amsthm}
\usepackage{float}
\usepackage{authblk}
\usepackage{lscape}
\usepackage{listings}
\usepackage{inconsolata} 

\usepackage{makecell} 
\definecolor{PastelGreen}{HTML}{8DC73F}
\definecolor{DarkForestGreen}{HTML}{316A59}
\definecolor{DarkOrange}{HTML}{FF8C00}
\definecolor{CodeGray}{gray}{0.98}

\lstdefinestyle{textarena}{
    backgroundcolor=\color{CodeGray},
    commentstyle=\color{DarkForestGreen},
    keywordstyle=\color{DarkOrange}\bfseries,
    stringstyle=\color{DarkForestGreen},
    basicstyle=\ttfamily\small,
    frame=single,
    showstringspaces=false,
    breaklines=true,
    tabsize=4,
    keepspaces=true,
    language=Python,
}

\usepackage{natbib}
\usepackage{placeins}

\usepackage[many]{tcolorbox}
\usepackage{enumitem}
\usepackage{wrapfig}

\newtcolorbox{codebox}{
    enhanced,
    colback=white,
    colframe=gray!20,
    boxrule=0.5pt,
    arc=3pt,
    top=3pt,
    bottom=3pt,
    left=4pt,
    right=4pt,
    boxsep=2pt,
    before skip=8pt,
    after skip=8pt
}

\definecolor{darkforestgreen}{RGB}{142, 182, 155} 
\hypersetup{
    colorlinks,
    linkcolor={darkforestgreen},
    citecolor={darkforestgreen},
    urlcolor={darkforestgreen}
}

\newcommand{\cmark}{\textcolor{green}{\ding{51}}}%
\newcommand{\xmark}{\textcolor{red}{\ding{55}}}%

\usepackage{skak} 

\title{TextArena}
\author[$\pawn$,*]{Leon Guertler}
\author[$\knight$,†]{Bobby Cheng}
\author[$\bishop$]{Simon Yu} 
\author[$\rook$]{Bo Liu}
\author[$\queen$]{Leshem Choshen}
\author[$\pawn$,$\knight$]{Cheston Tan}

\affil[ ]{$\pawn$ Centre for Frontier AI Research (CFAR), A*STAR}
\affil[ ]{$\knight$ Institute of High Performance Computing, A*STAR}
\affil[ ]{$\bishop$ Northeastern University}
\affil[ ]{$\rook$ National University of Singapore}
\affil[ ]{$\queen$ MIT, MIT-IBM Watson AI Lab}

\usepackage{fontawesome5} 
\usepackage{hyperref}    

\affil[ ]{}
\vspace{-18pt}
\affil[  ]{{\faPlayCircle} \textbf{Play:} \url{https://www.textarena.ai/}}
\affil[ ]{{\faTrophy} \textbf{Leaderboard:} \url{https://www.textarena.ai/leaderboard}}
\affil[ ]{{\faGithub} \textbf{Code:} \url{https://github.com/LeonGuertler/TextArena}}

\begin{document}

\maketitle
\renewcommand{\thefootnote}{\fnsymbol{footnote}}
\footnotetext[1]{Corresponding author: \texttt{Guertlerlo@cfar.a-star.edu.sg}}
\footnotetext[2]{Corresponding author: \texttt{chengxy@i2r.a-star.edu.sg}}

\renewcommand{\thefootnote}{\arabic{footnote}}

\vspace{-55pt}
\begin{figure}[tbh]
    \centering
    \includegraphics[width=0.50\linewidth]{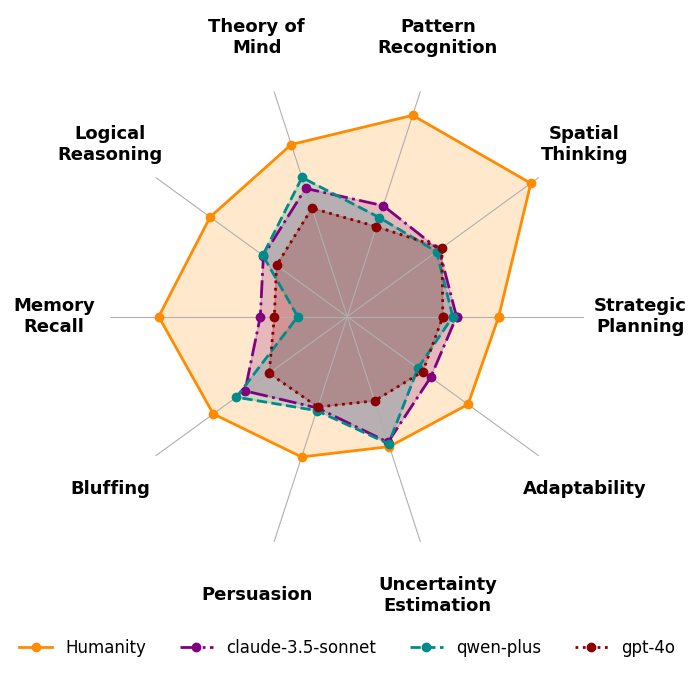}
    \caption{TextArena Soft-skill comparison. Frontier models and Humanity are compared across ten key skills. Each skill is normalised separately for presentation; see the leaderboard for full data.}
    \label{fig:skill-comparison}
\end{figure}

\begin{abstract}
TextArena is an open-source collection of competitive text-based games for training and evaluation of agentic behavior in Large Language Models (LLMs). It spans 57+ unique environments (including single-player, two-player, and multi-player setups) and allows for easy evaluation of model capabilities via an \textbf{online-play} system (against humans and other submitted models) with real-time TrueSkill\texttrademark{} scores. Traditional benchmarks rarely assess dynamic social skills such as negotiation, theory of mind, and deception, creating a gap that TextArena addresses. Designed with research, community and extensibility in mind, TextArena emphasizes ease of adding new games, adapting the framework, testing models, playing against the models, and training models. Detailed documentation of environments, games, leaderboard, and examples are available on \href{https://github.com/LeonGuertler/TextArena}{GitHub} and \href{https://www.textarena.ai/}{textarena.ai}. 
\end{abstract}
\begin{figure}[h!]
    \centering
    \includegraphics[width=.99\linewidth]{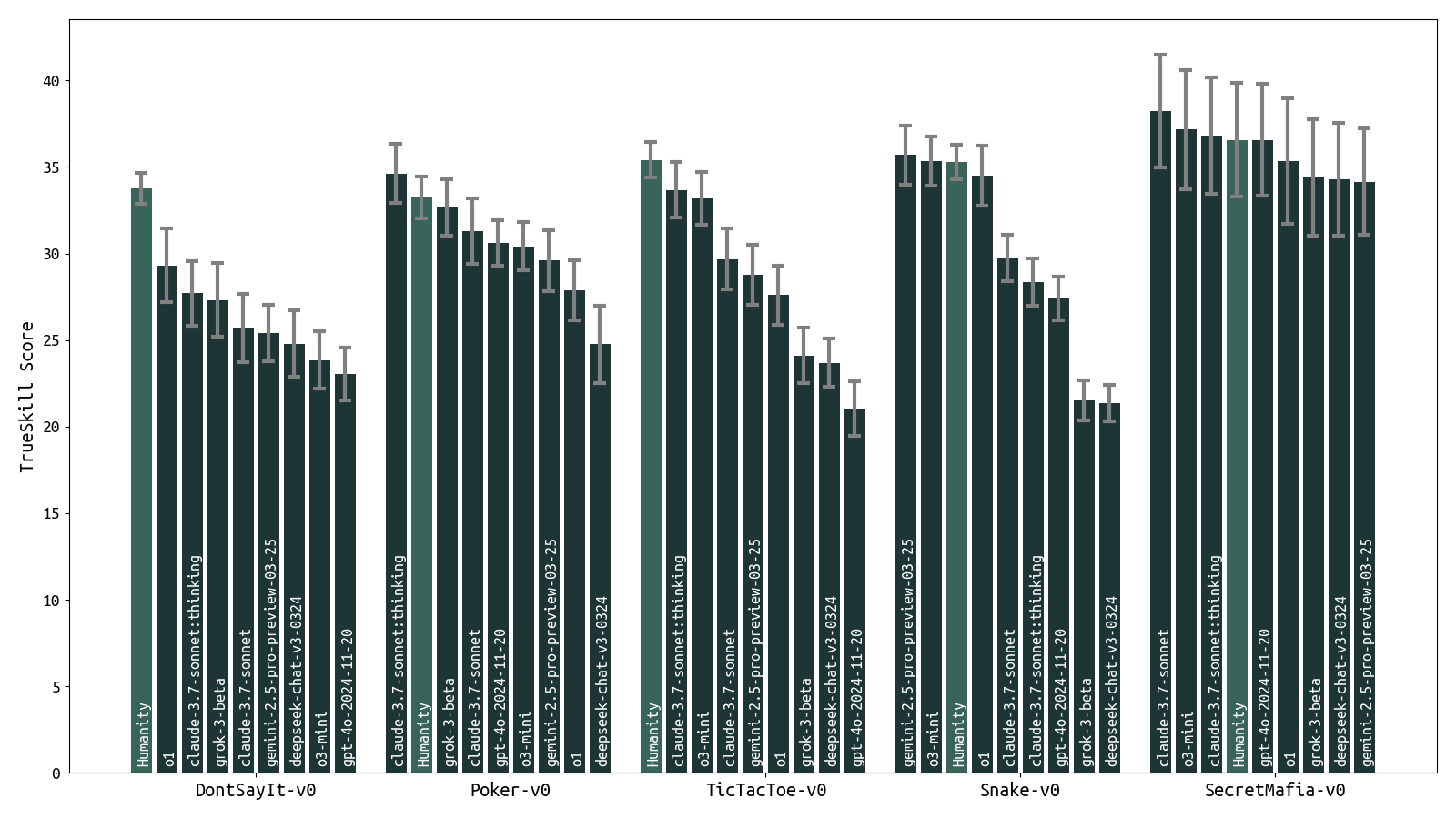}
    \caption{Preliminary model rankings for a subset of models and games. Game-play results are influenced by both the models' ability to play the games and their ability to understand the rules and format. For example, some reasoning models can sometimes reveal their cards or roles during game-play.}
    \label{fig:results}
\end{figure}


\section{Introduction}
\label{sec:introduction}


Scaling large language models has led to remarkable improvements in performance across various benchmarks. Models like GPT-4o~\citep{openai2024gpt4o}, Claude~\citep{anthropic2024claude35sonnet}, and Gemini~\citep{geminiteam2024gemini15unlockingmultimodal} have achieved near-perfect scores on traditional benchmarks like MMLU~\citep{mmlu} and HumanEval~\citep{humaneval}. Due to the recent progress on reasoning models (like OpenAI o1~\citet{openai2024openaio1card} and DeepSeek R1~\citet{deepseekai2025deepseekr1incentivizingreasoningcapability}), even more complex evaluations like the ARC-AGI challenge \citep{chollet2025arcprize2024technical} are approaching saturation, suggesting the need for a new evaluation paradigm. 

Two ad-hoc solutions to this are extending existing benchmarks~\citep{white2024livebenchchallengingcontaminationfreellm, kiela2021dynabenchrethinkingbenchmarkingnlp} and coming up with ever harder benchmarks, like "Humanity's Last Exam"~\citep{phan2025hle}. However, as models keep improving, it is conceivable that soon it will be infeasible for humans to come up with new, more challenging benchmarks. 

We argue that a more sustainable alternative to these absolute measures of performance is a relative one. The advantage thereof is that there is no clear upper limit of performance that can be reached, and thus, as long as models differ in capabilities, a ranking can be achieved. Chatbot Arena~\citep{chiang2024chatbotarenaopenplatform} follows such a strategy; there, humans select which of two LLM-generated answers they prefer. However, we aim to further circumvent human costs and biases, particularly as models approach or exceed the skill level of domain experts, making it increasingly challenging for humans to judge answer quality effectively and at scale.

Thus, we present TextArena, a comprehensive framework for evaluating language models through \textbf{competitive gameplay}. The initial release encompasses 57+ diverse text-based games\footnote{As of publication, the collection has grown to 74 games and continues to expand.}, including single-player, two-player, and multi-player scenarios (see current list in App.~\ref{ap:games}).

These games test a wide \textbf{range of capabilities}, including theory of mind, persuasion, deception, spatial reasoning, long-term planning and other social skills that traditional benchmarks typically do not assess. TextArena facilitates \textbf{offline} model development/training and \textbf{online competition} between models and human players (both model vs model and model vs human), with performance tracked through a real-time TrueSkill\texttrademark{}~\citep{10.5555/2976456.2976528} leaderboard that provides dynamic, relative capability measurements.

Within TextArena, LLMs interact with the environment in a dynamic, challenging and measurable way. Each agent acts based on its understanding of the current state and opponent's behaviors. The environment provides observations and rewards, enabling models to refine their strategies over time. This framework, inspired by platforms like OpenAI Gym~\citep{DBLP:journals/corr/BrockmanCPSSTZ16}, which standardize reinforcement learning interactions, creates opportunities for models to develop and demonstrate complex reasoning, negotiation, and decision-making capabilities in dynamic scenarios.

The framework's design emphasizes accessibility and extensibility, inviting researchers to \textbf{contribute games} and additional evaluation scenarios. Especially verifiable games that separate a specific skill and provide a natural scenario where it manifests. This communal and collaborative approach aims to create a living framework that evolves alongside the advancing model capabilities, providing sustained value for assessing LLMs. 

Given the re-animated focus on reinforcement learning, following the release of DeepSeek-R1~\citep{deepseekai2025deepseekr1incentivizingreasoningcapability} it is worth emphasizing that TextArena can serve as a source of near infinite training data for reinforcement learning with a dynamic curriculum of difficulty (via self-play), and thus in addition to improving soft skills like long-term planning, negotiation, theory of mind or deception, it is conceivable that this may serve as a further scaling paradigm for multi-turn, agentic reasoning models.

Overall, TextArena presents a versatile set of resources for interactive text games:
\begin{enumerate}
    \item A unified framework to describe games between models and a Gym-like framework suitable for RL training.
    \item 57+ games implemented in this framework.
    \item UI for humans to play against the models, supporting any games added.
    \item Leaderboards to compare general models, dedicated models and humans.
    \item Community support to using, playing, adding models, and further research with TextArena.
\end{enumerate}

\section{Design Choices}
\label{sec:design_choices}
In developing TextArena, our primary objectives were ease of adoption, use and extension. To address the former two, we kept the code interfaces used as similar to OpenAI Gym (aka Gymnasium)~\citep{DBLP:journals/corr/BrockmanCPSSTZ16} as possible and adopted their philosophy of stack-able wrappers. This design choice makes TextArena particularly well-suited for RL, providing researchers with a unified interface to diverse text-based environments. To further improve extensibility, we streamlined much of the shared game functionalities, making it easy and fast to add new environments to TextArena. 

To highlight these design choices, below is a short example script, showing how to use TextArena (See App.~\ref{app:online_play} for an example of how to add a model to play online).

For more detailed documentation, as well as tutorials for training and evaluation of models, check out \href{https://www.textarena.ai/}{textarena.ai}.

\begin{lstlisting}%[language=Python]
import textarena as ta

# Initialize agents
agents = {
    0: ta.agents.OpenRouterAgent(model_name="GPT-4o-mini"),
    1: ta.agents.OpenRouterAgent(model_name="anthropic/claude-3.5-haiku"),
}

# Initialize environment
env = ta.make(env_id=["TicTacToe-v0", "SpellingBee-v0"])
env = ta.wrappers.LLMObservationWrapper(env=env)

env.reset(num_players=len(agents))
done = False
while not done:
    player_id, observation = env.get_observation()
    action = agents[player_id](observation)
    done, info = env.step(action=action)
rewards = env.close()
\end{lstlisting}

\section{Environments}
\label{sec:environments}
To provide a rich and diverse training ground and evaluation set for the models, we have so far created 57+ text-based games, including original games, slightly adjusted games and novel games; for single-, two and multi-player setups. The environments cover a large range of hard and soft skills, including: Reasoning, Theory of Mind, Risk Assessment, Vocabulary Skills, Pattern Recognition, Spatial Reasoning, Planning, Memory, Deception, Negotiation, Persuasion, Resource Management, and many more. 

Importantly, all games used are either naively text-based or adapted to be text-based. 

A comprehensive, up-to-date, list of the environments, as well as additional information and documentation for each, can be found on \href{https://github.com/LeonGuertler/TextArena}{GitHub}. 

\begin{figure}
    \centering
    \includegraphics[width=.99\linewidth]{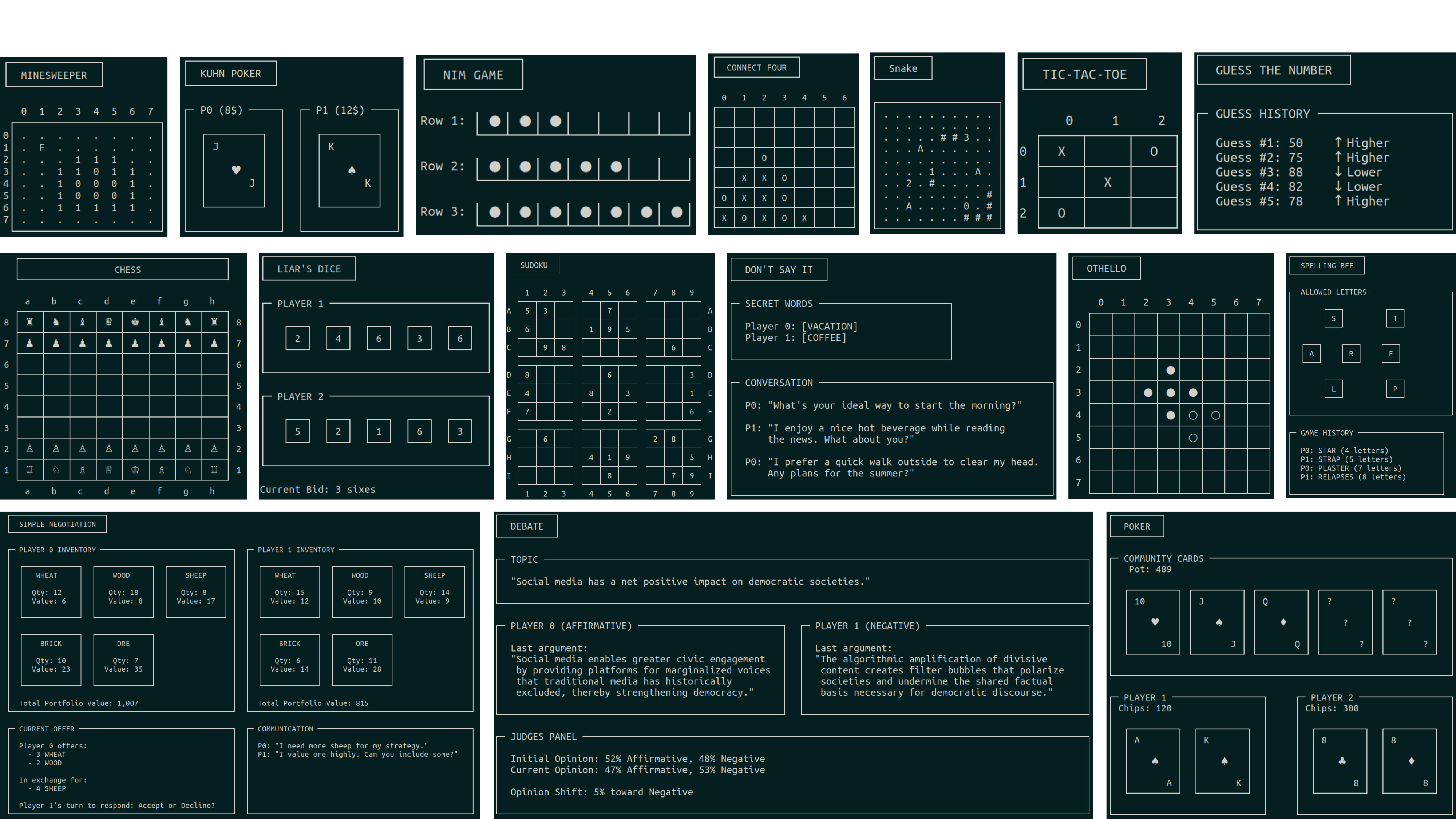}
    \caption{Images of some (rendered) TextArena environments.}
    \label{fig:envs}
\end{figure}

\section{Online Evaluation}
\label{sec:online}
TextArena employs a dynamic, competitive-based evaluation system to assess model performances of frontier models, community submitted models and humans as a baseline. We match, score and share a leaderboard and statistics on those models. The online leaderboard tracks performance through TrueSkill\texttrademark{}~\citep{10.5555/2976456.2976528}, a bayesian skill rating system originally developed for matchmaking in competitive games. This rating system is particularly well-suited for TextArena as it:
\begin{enumerate}
    \item Accurately rates players in both team-based and individual competitions
    \item Handles matches with varying numbers of players
    \item In our experiments, consistently converged faster to a reliable skill estimate than the traditional Elo system
    \item Appropriately manages uncertainty for new participants
\end{enumerate}

Each model is initialized with a TrueSkill\texttrademark{} rating ($\mu=25$, $\sigma=\frac{25}{3}$), with ratings adjusted after every match. Human players are collectively represented as "Humanity" on the leaderboard, providing a natural benchmark against which to measure model performance. This approach enables direct comparison between different models and between models and human players, creating a comprehensive ranking system that evolves as participants' abilities change over time.

Beyond overall performance, TextArena provides deeper insight into model capabilities through soft-skill profiling. Each environment is tagged with up to five soft skills (\textit{Strategic Planning}, \textit{Spatial Thinking}, \textit{Pattern Recognition}, \textit{Theory of Mind}, \textit{Logical Reasoning}, \textit{Memory Recall}, \textit{Bluffing}, \textit{Persuasion}, \textit{Uncertainty Estimation}, and \textit{Adaptability}) with corresponding weights. As models accumulate ratings across multiple environments, their aptitude in each skill category is estimated by calculating the weighted average of relevant environment scores.

This granular evaluation reveals specific strengths and weaknesses across models (Figure \ref{fig:skill-comparison}), providing researchers with actionable insights beyond overall rankings. For instance, while two models might achieve similar aggregate scores, one might excel at \textit{Uncertainty Estimation} and \textit{Bluffing} while the other demonstrates superior \textit{Persuasion} capabilities.

The online competition system facilitates both \textbf{Model vs Model} and \textbf{Model vs Human} play across our diverse library of environments. This multi-faceted evaluation approach provides a more nuanced understanding of model capabilities compared to static benchmarks. This is particularly relevant for assessing social skills like negotiation, deception, and theory of mind.

While we impose no formal restrictions on model submissions, we encourage researchers to submit different model variants under distinct names to maintain clarity in the leaderboard. So far, we have evaluated 283 models online, including community submissions and the 64 official models hosted by the platform. This openness supports our goal of creating a collaborative community around TextArena while still providing meaningful comparative evaluations\footnote{Live leaderboard: \url{https://www.textarena.ai/leaderboard}}.


\section{Related Work}
\vspace{-1.5em}
\begin{table}[tbhp]
    \centering
    \vspace{8pt}
    \resizebox{\textwidth}{!}{
    \begin{tabular}{lccc|cccc}
        \toprule
        \multirow{2}{*}{\textbf{Benchmark/Study}} & \multicolumn{3}{c|}{\textbf{Number of Environments}} & \multirow{2}{*}{\textbf{Gym-Compatible API}} & \multirow{2}{*}{\textbf{Online Evaluation}} & \multirow{2}{*}{\textbf{Model vs Model}} & \multirow{2}{*}{\textbf{Model vs Human}} \\
        \cmidrule(lr){2-4}
        & \textbf{Single-Player} & \textbf{Two-Player} & \textbf{Multi-Player} & & & & \\
        \midrule
        \textbf{Clembench}~\citep{chalamalasetti2023clembench} & 0 & 5 & 0 & \xmark & \xmark & \cmark & \xmark \\
        \textbf{LMRL-Gym}~\citep{abdulhai2023lmrl} & 5 & 3 & 0 & \cmark & \xmark & \cmark & \xmark \\
        \textbf{GameBench}~\citep{costarelli2024gamebench} & 0 & 3 & 6 & \cmark & \xmark & \cmark & \cmark \\
        \textbf{Game-theoretic LLM}~\citep{hua2024gametheoretic} & 0 & 11 & 0 & \xmark & \xmark & \cmark & \xmark \\
        \textbf{LAMEN}~\citep{davidson2024evaluating} & 0 & 6 & 0 & \xmark & \xmark & \cmark & \xmark \\
        \textbf{GTBench}~\citep{duan2024gtbench} & 0 & 10 & 0 & \xmark & \xmark & \cmark & \xmark \\
        \textbf{GameArena}~\citep{hu2024gamearena} & 0 & 3 & 0 & \xmark & \xmark & \xmark & \cmark \\
        \textbf{SPIN-Bench}~\citep{yao2025spinbench} & 21 & 3 & 2 & \cmark & \xmark & \cmark & \cmark \\
        \midrule
        \textbf{TextArena (Ours)} & 16 & 47 & 11 & \cmark & \cmark & \cmark & \cmark \\
        \bottomrule
    \end{tabular}
    }
    \vspace{8pt}
    \caption{Comparison of recent benchmarks for evaluating large language models (LLMs) in game-based interaction scenarios. The table summarizes the number of supported environments (Single-Player, Two-Player, Multi-Player), Gym-compatible API availability, online evaluation support, and capabilities for model-vs-model and model-vs-human interaction.}
    \label{tab:related_work_comparison}
\end{table}

Table~\ref{tab:related_work_comparison} provides a comprehensive comparison of recent benchmarks for evaluating large language models (LLMs) in game-based interaction scenarios. We analyze these frameworks across seven key dimensions: the number of environments in three categories (single-player, two-player, and multi-player games) and four technical capabilities (Gym-compatible API, online evaluation support, model versus model evaluation, and model versus human evaluation).

Existing benchmarks exhibit varied strengths and limitations. \textbf{Clembench}~\citep{chalamalasetti2023clembench} offers five two-player text-based environments with model versus model capabilities but lacks Gym compatibility and human evaluation features. \textbf{LMRL-Gym}~\citep{abdulhai2023lmrl} provides five single-player and three two-player environments with a Gym-compatible API, though it lacks human evaluation support. \textbf{GameBench}~\citep{costarelli2024gamebench} offers greater environment diversity with three two-player and six multi-player games, supporting both Gym compatibility and human evaluation capabilities.

More specialized frameworks include \textbf{Game-theoretic LLM}~\citep{hua2024gametheoretic} with eleven two-player environments and \textbf{LAMEN}~\citep{davidson2024evaluating} with six two-player environments, both supporting model versus model evaluation but lacking other technical capabilities. \textbf{GTBench}~\citep{duan2024gtbench} concentrates on ten two-player game-theoretic environments but offers limited technical features. \textbf{GameArena}~\citep{hu2024gamearena} specializes in three two-player environments with human evaluation as its distinguishing feature. \textbf{SPIN-Bench}~\citep{yao2025spinbench} provides a balanced distribution with one single-player, three two-player, and two multi-player environments, offering Gym compatibility and model versus model evaluation.

\textbf{TextArena} (our proposed benchmark) addresses these limitations by providing the most comprehensive coverage across all dimensions. With 16 single-player, 47 two-player, and 11 multi-player environments, TextArena offers substantially greater variety than existing benchmarks. It is the only framework that fully supports all four technical capabilities, enabling flexible evaluation across different interaction scenarios, reinforcement learning applications, and both model-to-model and model-to-human evaluations within a unified, extensible platform.

\section{Future Directions}
In the future, we hope to extend TextArena in several ways.
\begin{itemize}
    \item \textit{RL training}: Training reasoning models on game environments. We believe this would lead to the next training paradigm, serving as a new kind of data source.
    \item \textit{Public engagement}: We invite researchers and enthusiasts to contribute to TextArena by collaborating on research, adding games, testing models and playing against LLMs. We created a \href{https://discord.com/invite/KMndsqwMaZ}{Discord} channel to foster research collaborations. To encourage user engagement, we host 64 state-of-the-art models available to play online for free.
    \item \textit{Data Release}: We will release datasets including game-play trajectories between humans and models, such as OpenAI o1, Claude-3.7-Sonnet, and Gemini-2.5-Pro, to facilitate further research.
    \item \textit{VideoGameArena}: Building on our work with competitive text-based games, we aim to benchmark models in competitive frame-based environments, where agents would compete in real time using directional and key-based inputs.
\end{itemize}

\section*{Acknowledgements}
We thank Simone Romeo for the many great game suggestions; Ananya Balehithlu, Ayudh Saxena, Romir Patel, and Vincent Cheng for contributing environments; Henry Mao and Gabriel Chua for making TextArena MCP-compatible; Dylan Hillier for contributing to the code; Weiyan Shi for supporting the ideas; and OpenRouter, Anthropic, and AWS for supporting TextArena in various capacities.

\bibliographystyle{iclr}  
\bibliography{references}  

\newpage
\appendix

\section{Covered Games}\label{ap:games}
\begin{table}[h]
\centering
\small
\resizebox{\textwidth}{!}{
\begin{tabular}{lccccccccccc}
\toprule
\textbf{Game Name} & \textbf{Players} & \rotatebox{45}{{Strat.}} & \rotatebox{45}{{Spatial}} & \rotatebox{45}{{Pattern}} & \rotatebox{45}{{ToM}} & \rotatebox{45}{{Logic}} & \rotatebox{45}{{Mem.}} & \rotatebox{45}{{Bluff}} & \rotatebox{45}{{Pers.}} & \rotatebox{45}{{Uncert.}} & \rotatebox{45}{{Adapt.}} \\
\midrule
CarPuzzle$^{*}$             & 1 & \ding{109} & \ding{108} & \ding{108} & \ding{109} & \ding{108} & \ding{109} & \ding{109} & \ding{109} & \ding{109} & \ding{109} \\
Crosswords            & 1 & \ding{109} & \ding{109} & \ding{108} & \ding{109} & \ding{108} & \ding{108} & \ding{109} & \ding{109} & \ding{109} & \ding{109} \\
FifteenPuzzle         & 1 & \ding{109} & \ding{108} & \ding{108} & \ding{109} & \ding{108} & \ding{109} & \ding{109} & \ding{109} & \ding{109} & \ding{109} \\
GuessTheNumber        & 1 & \ding{109} & \ding{109} & \ding{108} & \ding{109} & \ding{108} & \ding{109} & \ding{109} & \ding{109} & \ding{108} & \ding{109} \\
GuessWho              & 1 & \ding{109} & \ding{109} & \ding{109} & \ding{108} & \ding{108} & \ding{108} & \ding{109} & \ding{109} & \ding{108} & \ding{108} \\
Hangman               & 1 & \ding{109} & \ding{109} & \ding{109} & \ding{109} & \ding{108} & \ding{108} & \ding{109} & \ding{109} & \ding{109} & \ding{108} \\
LogicPuzzle           & 1 & \ding{108} & \ding{109} & \ding{108} & \ding{109} & \ding{108} & \ding{108} & \ding{109} & \ding{109} & \ding{109} & \ding{108} \\
Mastermind            & 1 & \ding{108} & \ding{109} & \ding{108} & \ding{109} & \ding{108} & \ding{109} & \ding{109} & \ding{109} & \ding{109} & \ding{109} \\
MathProof$^{*}$             & 1 & \ding{109} & \ding{109} & \ding{108} & \ding{109} & \ding{108} & \ding{108} & \ding{109} & \ding{109} & \ding{109} & \ding{109} \\
Minesweeper           & 1 & \ding{109} & \ding{109} & \ding{108} & \ding{109} & \ding{108} & \ding{109} & \ding{109} & \ding{109} & \ding{108} & \ding{109} \\
Sudoku                & 1 & \ding{109} & \ding{109} & \ding{108} & \ding{109} & \ding{108} & \ding{108} & \ding{109} & \ding{109} & \ding{109} & \ding{109} \\
TowerOfHanoi          & 1 & \ding{108} & \ding{109} & \ding{109} & \ding{109} & \ding{108} & \ding{109} & \ding{109} & \ding{109} & \ding{109} & \ding{109} \\
TwentyQuestions       & 1 & \ding{109} & \ding{109} & \ding{109} & \ding{108} & \ding{108} & \ding{109} & \ding{109} & \ding{109} & \ding{108} & \ding{109} \\
WordLadder            & 1 & \ding{109} & \ding{109} & \ding{108} & \ding{109} & \ding{108} & \ding{108} & \ding{109} & \ding{109} & \ding{109} & \ding{109} \\
WordSearch            & 1 & \ding{109} & \ding{109} & \ding{108} & \ding{109} & \ding{108} & \ding{108} & \ding{109} & \ding{109} & \ding{109} & \ding{109} \\
Wordle                & 1 & \ding{109} & \ding{109} & \ding{108} & \ding{109} & \ding{108} & \ding{108} & \ding{109} & \ding{109} & \ding{109} & \ding{109} \\

\midrule
AirLandAndSea$^{\text{†,*}}$       & 2 & \ding{108} & \ding{109} & \ding{109} & \ding{108} & \ding{108} & \ding{109} & \ding{109} & \ding{109} & \ding{108} & \ding{108} \\
BattleOfSexes$^{\text{†,*}}$       & 2 & \ding{108} & \ding{109} & \ding{109} & \ding{108} & \ding{109} & \ding{109} & \ding{109} & \ding{108} & \ding{109} & \ding{109} \\
Battleship            & 2 & \ding{109} & \ding{108} & \ding{108} & \ding{109} & \ding{108} & \ding{109} & \ding{109} & \ding{109} & \ding{108} & \ding{109} \\
Brass$^{*}$                 & 2 & \ding{108} & \ding{109} & \ding{109} & \ding{108} & \ding{109} & \ding{109} & \ding{109} & \ding{109} & \ding{108} & \ding{109} \\
Breakthrough$^{\text{¶}}$        & 2 & \ding{108} & \ding{108} & \ding{108} & \ding{109} & \ding{108} & \ding{109} & \ding{109} & \ding{109} & \ding{109} & \ding{109} \\
Checkers              & 2 & \ding{108} & \ding{109} & \ding{108} & \ding{109} & \ding{108} & \ding{109} & \ding{109} & \ding{109} & \ding{109} & \ding{109} \\
Chess                 & 2 & \ding{108} & \ding{108} & \ding{108} & \ding{109} & \ding{108} & \ding{108} & \ding{109} & \ding{109} & \ding{109} & \ding{109} \\
ConnectFour           & 2 & \ding{108} & \ding{108} & \ding{108} & \ding{109} & \ding{108} & \ding{109} & \ding{109} & \ding{109} & \ding{109} & \ding{109} \\
Debate                & 2 & \ding{109} & \ding{109} & \ding{109} & \ding{108} & \ding{108} & \ding{109} & \ding{109} & \ding{108} & \ding{109} & \ding{108} \\
DontSayIt             & 2 & \ding{109} & \ding{109} & \ding{109} & \ding{108} & \ding{109} & \ding{108} & \ding{108} & \ding{109} & \ding{109} & \ding{108} \\
DracoGame$^{\text{†,*}}$           & 2 & \ding{108} & \ding{109} & \ding{109} & \ding{108} & \ding{109} & \ding{108} & \ding{109} & \ding{109} & \ding{109} & \ding{109} \\
DuopolisticCompetition$^{\text{†,*}}$ & 2 & \ding{108} & \ding{109} & \ding{109} & \ding{108} & \ding{109} & \ding{109} & \ding{108} & \ding{109} & \ding{109} & \ding{109} \\
EscalationGame$^{\text{†,*}}$      & 2 & \ding{109} & \ding{109} & \ding{109} & \ding{108} & \ding{109} & \ding{109} & \ding{108} & \ding{109} & \ding{108} & \ding{109} \\
Hive$^{\text{†,*}}$                & 2 & \ding{108} & \ding{108} & \ding{108} & \ding{109} & \ding{108} & \ding{109} & \ding{109} & \ding{109} & \ding{109} & \ding{109} \\
HotColdGame$^{\text{†,*}}$         & 2 & \ding{109} & \ding{109} & \ding{109} & \ding{109} & \ding{108} & \ding{109} & \ding{109} & \ding{109} & \ding{108} & \ding{108} \\
IntegrativeDistributiveNegotiation$^{\text{§,*}}$ & 2 & \ding{108} & \ding{109} & \ding{109} & \ding{108} & \ding{108} & \ding{109} & \ding{109} & \ding{108} & \ding{109} & \ding{108} \\
IteratedPrisonersDilemma & 2 & \ding{108} & \ding{109} & \ding{109} & \ding{108} & \ding{109} & \ding{109} & \ding{109} & \ding{109} & \ding{108} & \ding{108} \\
Jaipur$^{*}$                & 2 & \ding{108} & \ding{109} & \ding{109} & \ding{108} & \ding{109} & \ding{108} & \ding{108} & \ding{108} & \ding{108} & \ding{109} \\
KuhnPoker$^{\text{¶}}$           & 2 & \ding{108} & \ding{109} & \ding{109} & \ding{108} & \ding{109} & \ding{109} & \ding{108} & \ding{109} & \ding{108} & \ding{109} \\
LetterAuction         & 2 & \ding{108} & \ding{109} & \ding{109} & \ding{109} & \ding{109} & \ding{109} & \ding{108} & \ding{108} & \ding{108} & \ding{109} \\
MemoryGame            & 2 & \ding{109} & \ding{109} & \ding{109} & \ding{109} & \ding{109} & \ding{108} & \ding{109} & \ding{109} & \ding{109} & \ding{109} \\
MonopolyGame$^{\text{†,*}}$        & 2 & \ding{108} & \ding{109} & \ding{109} & \ding{108} & \ding{109} & \ding{109} & \ding{108} & \ding{108} & \ding{108} & \ding{108} \\
Nim$^{\text{¶}}$                 & 2 & \ding{108} & \ding{109} & \ding{109} & \ding{109} & \ding{108} & \ding{109} & \ding{109} & \ding{109} & \ding{109} & \ding{109} \\
Othello (Reversi)     & 2 & \ding{108} & \ding{109} & \ding{108} & \ding{109} & \ding{108} & \ding{109} & \ding{109} & \ding{109} & \ding{109} & \ding{109} \\
PigDice$^{\text{¶}}$             & 2 & \ding{108} & \ding{109} & \ding{109} & \ding{109} & \ding{108} & \ding{109} & \ding{109} & \ding{109} & \ding{108} & \ding{109} \\
PrisonersDilemma$^{\text{†}}$    & 2 & \ding{108} & \ding{109} & \ding{109} & \ding{108} & \ding{109} & \ding{109} & \ding{109} & \ding{109} & \ding{108} & \ding{108} \\
Santorini$^{\text{†,*}}$           & 2 & \ding{108} & \ding{108} & \ding{109} & \ding{109} & \ding{108} & \ding{109} & \ding{109} & \ding{109} & \ding{109} & \ding{109} \\
ScenarioPlanning      & 2 & \ding{108} & \ding{109} & \ding{109} & \ding{108} & \ding{108} & \ding{109} & \ding{109} & \ding{108} & \ding{109} & \ding{109} \\
SeaBattle$^{\text{†,*}}$           & 2 & \ding{109} & \ding{108} & \ding{108} & \ding{109} & \ding{108} & \ding{109} & \ding{109} & \ding{109} & \ding{108} & \ding{109} \\
SimpleBlindAuction$^{\text{¶}}$  & 2 & \ding{109} & \ding{109} & \ding{109} & \ding{108} & \ding{109} & \ding{109} & \ding{108} & \ding{108} & \ding{108} & \ding{109} \\
SimpleNegotiation     & 2 & \ding{108} & \ding{109} & \ding{109} & \ding{108} & \ding{109} & \ding{109} & \ding{108} & \ding{109} & \ding{109} & \ding{108} \\
SpellingBee           & 2 & \ding{109} & \ding{109} & \ding{108} & \ding{109} & \ding{108} & \ding{108} & \ding{109} & \ding{109} & \ding{109} & \ding{108} \\
SpiteAndMalice        & 2 & \ding{108} & \ding{109} & \ding{109} & \ding{108} & \ding{108} & \ding{109} & \ding{109} & \ding{109} & \ding{108} & \ding{108} \\
StagHunt$^{\text{†,*}}$            & 2 & \ding{108} & \ding{109} & \ding{109} & \ding{108} & \ding{109} & \ding{109} & \ding{109} & \ding{109} & \ding{108} & \ding{109} \\
Stratego              & 2 & \ding{108} & \ding{109} & \ding{108} & \ding{108} & \ding{109} & \ding{109} & \ding{109} & \ding{109} & \ding{108} & \ding{108} \\
Taboo                 & 2 & \ding{108} & \ding{109} & \ding{109} & \ding{108} & \ding{108} & \ding{109} & \ding{109} & \ding{109} & \ding{109} & \ding{108} \\
Tak                   & 2 & \ding{108} & \ding{108} & \ding{108} & \ding{109} & \ding{108} & \ding{109} & \ding{109} & \ding{109} & \ding{109} & \ding{109} \\
TicTacToe             & 2 & \ding{108} & \ding{109} & \ding{109} & \ding{109} & \ding{108} & \ding{109} & \ding{109} & \ding{109} & \ding{109} & \ding{109} \\
TriGame$^{\text{†,*}}$             & 2 & \ding{108} & \ding{109} & \ding{109} & \ding{108} & \ding{108} & \ding{109} & \ding{109} & \ding{109} & \ding{109} & \ding{109} \\
TruthAndDeception     & 2 & \ding{109} & \ding{109} & \ding{109} & \ding{108} & \ding{108} & \ding{109} & \ding{108} & \ding{108} & \ding{108} & \ding{109} \\
UltimateTicTacToe     & 2 & \ding{108} & \ding{108} & \ding{108} & \ding{109} & \ding{108} & \ding{109} & \ding{109} & \ding{109} & \ding{109} & \ding{109} \\
WaitGoGame$^{\text{†,*}}$          & 2 & \ding{108} & \ding{109} & \ding{109} & \ding{109} & \ding{108} & \ding{109} & \ding{109} & \ding{109} & \ding{109} & \ding{108} \\
WordChains            & 2 & \ding{109} & \ding{109} & \ding{108} & \ding{109} & \ding{108} & \ding{108} & \ding{109} & \ding{109} & \ding{109} & \ding{108} \\
\bottomrule
\end{tabular}
}
\caption{
This table categorizes \textbf{single}-player and \textbf{two}-player games by the number of players and the primary cognitive or strategic skills emphasized in each game. Filled circles (\ding{108}) are skills that are relevant to each game, while empty circles (\ding{109}) indicate skills not emphasized. Games marked with † are drawn from \citet{costarelli2024gamebench}, ‡ from \citet{hua2024gametheoretic}, § from \citet{davidson2024evaluating}, and ¶ from \citet{duan2024gtbench}. \textit{Strat.} = Strategic Planning, \textit{Spatial} = Spatial Thinking, \textit{Pattern} = Pattern Recognition, \textit{ToM} = Theory of Mind, \textit{Logic} = Logical Reasoning, \textit{Mem.} = Memory Recall, \textit{Bluff} = Bluffing, \textit{Pers.} = Persuasion, \textit{Uncert.} = Uncertainty Estimation, \textit{Adapt.} = Adaptability. Games marked with ${*}$ have not been fully implemented yet.}
\end{table}

\subsection{Single-Player Games}
\begin{itemize}
    \item \textbf{CarPuzzle} A challenging puzzle game where players must rearrange vehicles in a grid to clear a path for the target car. \\
    \textbf{Skills:} \textit{Spatial Thinking} • \textit{Pattern Recognition} • \textit{Logical Reasoning}
    
    \item \textbf{Crosswords} A classic word puzzle game where players solve clues to fill in intersecting words on a grid. \\
    \textbf{Skills:} \textit{Memory Recall} • \textit{Pattern Recognition} • \textit{Logical Reasoning}
    
    \item \textbf{FifteenPuzzle} A sliding tile puzzle where players move numbered tiles within a grid to restore them to their original order. \\
    \textbf{Skills:} \textit{Spatial Thinking} • \textit{Logical Reasoning} • \textit{Pattern Recognition}
    
    \item \textbf{GuessTheNumber} A game where players deduce a hidden number within a given range using numerical hints. \\
    \textbf{Skills:} \textit{Logical Reasoning} • \textit{Uncertainty Estimation} • \textit{Pattern Recognition}
    
    \item \textbf{GuessWho} A deduction-based game where players ask strategic yes-or-no questions to identify a hidden character from a pool of possibilities. \\
    \textbf{Skills:} \textit{Theory of Mind} • \textit{Logical Reasoning} • \textit{Uncertainty Estimation} • \textit{Memory Recall} • \textit{Adaptability}
    
    \item \textbf{Hangman} A word-guessing game where players try to uncover a hidden word by guessing letters, with limited attempts before the game ends. \\
    \textbf{Skills:} \textit{Memory Recall} • \textit{Logical Reasoning} • \textit{Adaptability}
    
    \item \textbf{LogicPuzzle} A collection of puzzles that require players to use deductive reasoning to solve problems, such as riddles or grid-based challenges. \\
    \textbf{Skills:} \textit{Logical Reasoning} • \textit{Pattern Recognition} • \textit{Memory Recall} • \textit{Strategic Planning} • \textit{Adaptability}
    
    \item \textbf{Mastermind} A code-breaking game where players deduce a hidden sequence of colors or numbers based on feedback. \\
    \textbf{Skills:} \textit{Pattern Recognition} • \textit{Logical Reasoning} • \textit{Strategic Planning}
    
    \item \textbf{MathProof} A unique game that challenges players to construct mathematical proofs to solve puzzles. \\
    \textbf{Skills:} \textit{Logical Reasoning} • \textit{Pattern Recognition} • \textit{Memory Recall}
    
    \item \textbf{Minesweeper} A grid-based game where players use numerical clues to identify and avoid hidden mines. \\
    \textbf{Skills:} \textit{Uncertainty Estimation} • \textit{Logical Reasoning} • \textit{Pattern Recognition}
    
    \item \textbf{Sudoku} A number-placement puzzle where players fill a 9x9 grid so that each row, column, and subgrid contains all digits from 1 to 9. \\
    \textbf{Skills:} \textit{Pattern Recognition} • \textit{Logical Reasoning} • \textit{Memory Recall}
    
    \item \textbf{TowerOfHanoi} A mathematical puzzle where players move a stack of disks between pegs according to specific rules. \\
    \textbf{Skills:} \textit{Strategic Planning} • \textit{Logical Reasoning}
    
    \item \textbf{TwentyQuestions} A guessing game where players identify an object by asking up to twenty yes-or-no questions. \\
    \textbf{Skills:} \textit{Theory of Mind} • \textit{Logical Reasoning} • \textit{Uncertainty Estimation}
    
    \item \textbf{WordLadder} A word puzzle where players transform one word into another by changing one letter at a time, with each step being a valid word. \\
    \textbf{Skills:} \textit{Memory Recall} • \textit{Pattern Recognition} • \textit{Logical Reasoning}
    
    \item \textbf{WordSearch} A puzzle game where players locate hidden words within a grid of letters. \\
    \textbf{Skills:} \textit{Pattern Recognition} • \textit{Memory Recall} • \textit{Logical Reasoning}
    
    \item \textbf{Wordle} A popular word-guessing game where players deduce a hidden five-letter word using letter placement feedback. \\
    \textbf{Skills:} \textit{Pattern Recognition} • \textit{Logical Reasoning} • \textit{Memory Recall}
\end{itemize}

\subsection{Two-Player Games}
\begin{itemize}
    \item \textbf{AirLandAndSea} A multi-domain strategy game involving forces on land, air, and sea.\\
    \textbf{Skills:} \textit{Strategic Planning} • \textit{Theory of Mind} • \textit{Logical Reasoning} • \textit{Uncertainty Estimation} • \textit{Adaptability}
    
    \item \textbf{BattleOfSexes} A coordination game exploring conflicting preferences between two players.\\
    \textbf{Skills:} \textit{Strategic Planning} • \textit{Theory of Mind} • \textit{Persuasion}
    
    \item \textbf{Battleship} A guessing game where players try to sink each other's naval fleets.\\
    \textbf{Skills:} \textit{Spatial Thinking} • \textit{Pattern Recognition} • \textit{Logical Reasoning} • \textit{Uncertainty Estimation}
    
    \item \textbf{Brass} An economic strategy game focusing on industrial development and network building.\\
    \textbf{Skills:} \textit{Strategic Planning} • \textit{Theory of Mind} • \textit{Uncertainty Estimation}
    
    \item \textbf{Breakthrough}
    A tactical abstract board game with simple rules and deep strategy.\\
    \textbf{Skills:} \textit{Strategic Planning} • \textit{Spatial Thinking} • \textit{Pattern Recognition} • \textit{Logical Reasoning}
    
    \item \textbf{Checkers}
    A classic board game featuring simple moves and jumps toward victory.\\
    \textbf{Skills:} \textit{Strategic Planning} • \textit{Pattern Recognition} • \textit{Logical Reasoning}
    
    \item \textbf{Chess}
    A timeless battle of wits and tactics between two opponents.\\
    \textbf{Skills:} \textit{Strategic Planning} • \textit{Spatial Thinking} • \textit{Pattern Recognition} • \textit{Logical Reasoning} • \textit{Memory}
    
    \item \textbf{ConnectFour}
    A vertical alignment game where players drop tokens to connect four in a row.\\
    \textbf{Skills:} \textit{Strategic Planning} • \textit{Spatial Thinking} • \textit{Pattern Recognition} • \textit{Logical Reasoning}
    
    \item \textbf{Debate}
    A competitive game involving argumentation and persuasive tactics.\\
    \textbf{Skills:} \textit{Theory of Mind} • \textit{Logical Reasoning} • \textit{Persuasion} • \textit{Adaptability}
    
    \item \textbf{DontSayIt}
    A word‐based game where avoiding forbidden terms is key to success.\\
    \textbf{Skills:} \textit{Theory of Mind} • \textit{Memory} • \textit{Bluffing} • \textit{Adaptability}
    
    \item \textbf{DracoGame}
    A thematic game involving magical challenges and strategic choices.\\
    \textbf{Skills:} \textit{Strategic Planning} • \textit{Theory of Mind} • \textit{Memory}
    
    \item \textbf{DuopolisticCompetition}
    A simulation game modeling market rivalry between two competing firms.\\
    \textbf{Skills:} \textit{Strategic Planning} • \textit{Theory of Mind} • \textit{Bluffing}
    
    \item \textbf{EscalationGame}
    A game that models competitive escalation in conflict scenarios.\\
    \textbf{Skills:} \textit{Theory of Mind} • \textit{Bluffing} • \textit{Uncertainty Estimation}
    
    \item \textbf{Hive}
    An abstract insect-themed game where players surround their opponent.\\
    \textbf{Skills:} \textit{Strategic Planning} • \textit{Spatial Thinking} • \textit{Pattern Recognition} • \textit{Logical Reasoning}
    
    \item \textbf{HotColdGame}
    A proximity-based guessing game where hints of “hot” or “cold” guide players.\\
    \textbf{Skills:} \textit{Logical Reasoning} • \textit{Uncertainty Estimation} • \textit{Adaptability}
    
    \item \textbf{IntegrativeDistributiveNegotiation}
    A negotiation simulation that blends integrative and distributive bargaining.\\
    \textbf{Skills:} \textit{Strategic Planning} • \textit{Theory of Mind} • \textit{Logical Reasoning} • \textit{Persuasion} • \textit{Adaptability}
    
    \item \textbf{IteratedPrisonersDilemma}
    A repeated social dilemma game highlighting conflict between individual and collective interests.\\
    \textbf{Skills:} \textit{Strategic Planning} • \textit{Theory of Mind} • \textit{Uncertainty Estimation} • \textit{Adaptability}
    
    \item \textbf{Jaipur}
    A fast-paced card game of trade and tactical resource management.\\
    \textbf{Skills:} \textit{Strategic Planning} • \textit{Theory of Mind} • \textit{Memory} • \textit{Bluffing} • \textit{Persuasion} • \textit{Uncertainty Estimation}
    
    \item \textbf{KuhnPoker}
    A simplified poker game that emphasizes strategic betting and incomplete information.\\
    \textbf{Skills:} \textit{Strategic Planning} • \textit{Theory of Mind} • \textit{Bluffing} • \textit{Uncertainty Estimation}
    
    \item \textbf{LetterAuction}
    A bidding game where players acquire letters to form words under pressure.\\
    \textbf{Skills:} \textit{Strategic Planning} • \textit{Bluffing} • \textit{Persuasion} • \textit{Uncertainty Estimation}
    
    \item \textbf{MemoryGame}
    A classic matching game testing recall through repeated turns.\\
    \textbf{Skills:} \textit{Memory}
    
    \item \textbf{MonopolyGame}
    A property-trading game simulating real estate and market competition.\\
    \textbf{Skills:} \textit{Strategic Planning} • \textit{Theory of Mind} • \textit{Bluffing} • \textit{Persuasion} • \textit{Uncertainty Estimation} • \textit{Adaptability}
    
    \item \textbf{Nim}
    A mathematical removal game that challenges logical strategy and planning.\\
    \textbf{Skills:} \textit{Strategic Planning} • \textit{Logical Reasoning}
    
    \item \textbf{Othello (Reversi)}
    A board game of flipping discs to dominate the board by the end of play.\\
    \textbf{Skills:} \textit{Strategic Planning} • \textit{Pattern Recognition}
    
    \item \textbf{PigDice}
    A dice game that tests risk management and probability with each roll.\\
    \textbf{Skills:} \textit{Strategic Planning} • \textit{Memory} • \textit{Uncertainty Estimation}
    
    \item \textbf{PrisonersDilemma}
    A classic strategic dilemma highlighting the tension between cooperation and self-interest.\\
    \textbf{Skills:} \textit{Strategic Planning} • \textit{Theory of Mind} • \textit{Uncertainty Estimation} • \textit{Adaptability}
    
    \item \textbf{Santorini}
    A visually appealing game where players build structures to outmaneuver their opponents.\\
    \textbf{Skills:} \textit{Strategic Planning} • \textit{Spatial Thinking} • \textit{Logical Reasoning}
    
    \item \textbf{ScenarioPlanning}
    A game centered on planning and decision-making under uncertainty.\\
    \textbf{Skills:} \textit{Strategic Planning} • \textit{Theory of Mind} • \textit{Logical Reasoning} • \textit{Persuasion}
    
    \item \textbf{SeaBattle}
    A naval combat game where players strategically hide and target enemy ships.\\
    \textbf{Skills:} \textit{Spatial Thinking} • \textit{Pattern Recognition} • \textit{Logical Reasoning} • \textit{Uncertainty Estimation}
    
    \item \textbf{SimpleBlindAuction}
    An auction game where bids are made without knowing the other players’ offers.\\
    \textbf{Skills:} \textit{Theory of Mind} • \textit{Bluffing} • \textit{Persuasion} • \textit{Uncertainty Estimation}
    
    \item \textbf{SimpleNegotiation}
    A straightforward bargaining game focusing on reaching mutually beneficial agreements.\\
    \textbf{Skills:} \textit{Strategic Planning} • \textit{Theory of Mind} • \textit{Bluffing} • \textit{Adaptability}
    
    \item \textbf{SpellingBee}
    A word challenge game where players race against time and pressure to spell words correctly.\\
    \textbf{Skills:} \textit{Pattern Recognition} • \textit{Logical Reasoning} • \textit{Memory} • \textit{Adaptability}
    
    \item \textbf{SpiteAndMalice}
    A competitive card game where players use strategy to thwart their opponents.\\
    \textbf{Skills:} \textit{Strategic Planning} • \textit{Theory of Mind} • \textit{Logical Reasoning} • \textit{Uncertainty Estimation} • \textit{Adaptability}
    
    \item \textbf{StagHunt}
    A game that explores cooperation and risk through a classic coordination dilemma.\\
    \textbf{Skills:} \textit{Strategic Planning} • \textit{Theory of Mind} • \textit{Uncertainty Estimation}
    
    \item \textbf{Stratego}
    A battle of wits where hidden pieces and surprise attacks decide the outcome.\\
    \textbf{Skills:} \textit{Strategic Planning} • \textit{Pattern Recognition} • \textit{Theory of Mind} • \textit{Uncertainty Estimation} • \textit{Adaptability}
    
    \item \textbf{Taboo}
    A fast-paced party game that challenges players to describe words without using forbidden terms.\\
    \textbf{Skills:} \textit{Strategic Planning} • \textit{Theory of Mind} • \textit{Logical Reasoning} • \textit{Adaptability}
    
    \item \textbf{Tak}
    An abstract strategy board game where players aim to create a road connecting opposite sides of the board.\\
    \textbf{Skills:} \textit{Strategic Planning} • \textit{Spatial Thinking} • \textit{Pattern Recognition} • \textit{Logical Reasoning}
    
    \item \textbf{TicTacToe}
    A simple grid game in which players try to align three marks in a row before their opponent.\\
    \textbf{Skills:} \textit{Strategic Planning} • \textit{Logical Reasoning}
    
    \item \textbf{TriGame}
    A strategic game that challenges players with triangular tactics and positional play.\\
    \textbf{Skills:} \textit{Strategic Planning} • \textit{Theory of Mind} • \textit{Logical Reasoning}
    
    \item \textbf{TruthAndDeception}
    A game of honesty and misdirection where players balance truth against lies.\\
    \textbf{Skills:} \textit{Theory of Mind} • \textit{Logical Reasoning} • \textit{Bluffing} • \textit{Persuasion} • \textit{Uncertainty Estimation}
    
    \item \textbf{UltimateTicTacToe}
    An advanced variant of Tic Tac Toe that introduces nested boards and greater strategic depth.\\
    \textbf{Skills:} \textit{Strategic Planning} • \textit{Spatial Thinking} • \textit{Pattern Recognition} • \textit{Logical Reasoning}
    
    \item \textbf{WaitGoGame}
    A game that blends waiting mechanics with strategic movement to challenge opponents.\\
    \textbf{Skills:} \textit{Strategic Planning} • \textit{Logical Reasoning} • \textit{Adaptability}
    
    \item \textbf{WordChains}
    A word association game where each answer must connect seamlessly to the next.\\
    \textbf{Skills:} \textit{Pattern Recognition} • \textit{Logical Reasoning} • \textit{Memory} • \textit{Adaptability}
\end{itemize}

\begin{table}[H]
\centering
\small
\resizebox{\textwidth}{!}{
\begin{tabular}{lccccccccccc}
\toprule
\textbf{Game Name} & \textbf{Players} & \rotatebox{45}{\textit{Strat.}} & \rotatebox{45}{\textit{Spatial}} & \rotatebox{45}{\textit{Pattern}} & \rotatebox{45}{\textit{ToM}} & \rotatebox{45}{\textit{Logic}} & \rotatebox{45}{\textit{Mem.}} & \rotatebox{45}{\textit{Bluff}} & \rotatebox{45}{\textit{Pers.}} & \rotatebox{45}{\textit{Uncert.}} & \rotatebox{45}{\textit{Adapt.}} \\
\midrule
Blind Auction       & 3–15& \ding{109} & \ding{109} & \ding{109} & \ding{108} & \ding{109} & \ding{109} & \ding{109} & \ding{108} & \ding{108} & \ding{109} \\ 
Character Conclave  & 3–15& \ding{108} & \ding{109} & \ding{109} & \ding{108} & \ding{109} & \ding{109} & \ding{109} & \ding{108} & \ding{109} & \ding{108} \\ 
Codenames$^{\text{†}}$ & 4  & \ding{108} & \ding{109} & \ding{108} & \ding{108} & \ding{108} & \ding{109} & \ding{109} & \ding{108} & \ding{109} & \ding{109} \\ 
Liar's Dice         & 2–15& \ding{109} & \ding{109} & \ding{109} & \ding{108} & \ding{109} & \ding{108} & \ding{108} & \ding{109} & \ding{108} & \ding{109} \\ 
Negotiation         & 3–15& \ding{108} & \ding{109} & \ding{109} & \ding{108} & \ding{109} & \ding{109} & \ding{109} & \ding{108} & \ding{109} & \ding{109} \\ 
Pit$^{\text{†,*}}$    & 3+  & \ding{109} & \ding{109} & \ding{109} & \ding{109} & \ding{109} & \ding{109} & \ding{109} & \ding{109} & \ding{108} & \ding{109} \\ 
Poker               & 2–15& \ding{108} & \ding{109} & \ding{109} & \ding{108} & \ding{109} & \ding{109} & \ding{108} & \ding{108} & \ding{108} & \ding{109} \\ 
Snake               & 2–15& \ding{108} & \ding{108} & \ding{109} & \ding{109} & \ding{109} & \ding{109} & \ding{109} & \ding{109} & \ding{109} & \ding{109} \\ 
Surround            & 2–15& \ding{108} & \ding{108} & \ding{109} & \ding{109} & \ding{109} & \ding{109} & \ding{109} & \ding{109} & \ding{109} & \ding{109} \\ 
Two Rooms and a Boom$^{\text{†}}$ & 6+ & \ding{108} & \ding{109} & \ding{109} & \ding{108} & \ding{109} & \ding{109} & \ding{109} & \ding{108} & \ding{109} & \ding{109} \\ 
Diplomacy           & 3–7 & \ding{108} & \ding{109} & \ding{109} & \ding{108} & \ding{109} & \ding{109} & \ding{109} & \ding{108} & \ding{109} & \ding{108} \\ 
SecretMafia & 5-15 & 
\ding{109} & \ding{109} & \ding{109} & \ding{108} & \ding{109} & \ding{108} & \ding{108} & \ding{108} & \ding{109} & \ding{108} \\
\bottomrule
\end{tabular}
}

\caption{This table categorizes \textbf{multi}-player games by the number of players and the primary cognitive or strategic skills emphasized in each game. Filled circles (\ding{108}) indicate which skills are relevant to each game, while empty circles (\ding{109}) indicate skills not emphasized. Games marked with special symbols are referenced from notable studies: games marked with † are drawn from \citet{costarelli2024gamebench}, ‡ from \citet{hua2024gametheoretic}, § from \citet{davidson2024evaluating}, and ¶ from \citet{duan2024gtbench}. \textit{Strat.} = Strategic Planning, \textit{Spatial} = Spatial Thinking, \textit{Pattern} = Pattern Recognition, \textit{ToM} = Theory of Mind, \textit{Logic} = Logical Reasoning, \textit{Mem.} = Memory Recall, \textit{Bluff} = Bluffing, \textit{Pers.} = Persuasion, \textit{Uncert.} = Uncertainty Estimation, \textit{Adapt.} = Adaptability. Games marked with ${*}$ have not been fully implemented yet.}
\label{tab:multi_player_games}
\end{table}

\subsection{Multi-Player Games}
\begin{itemize}
    \item \textbf{Blind Auction} A multi-player strategic auction game where players bid on items with different personal valuations. \\
    \textbf{Skills:} \textit{Persuasion} • \textit{Theory of Mind} • \textit{Uncertainty Estimation}
    
    \item \textbf{Character Conclave} A multi-player (3--15) text-based game where players engage in discussion with a limited character budget. \\
    \textbf{Skills:} \textit{Theory of Mind} • \textit{Persuasion} • \textit{Strategic Planning} • \textit{Adaptability}
    
    \item \textbf{Codenames} A team-based word association game requiring players to link related words. \\
    \textbf{Skills:} \textit{Theory of Mind} • \textit{Pattern Recognition} • \textit{Strategic Planning} • \textit{Logical Reasoning} • \textit{Persuasion}
    
    \item \textbf{Liar's Dice} A bluffing dice game where players make increasingly higher bids about the dice everyone has rolled while calling out suspected lies. \\
    \textbf{Skills:} \textit{Bluffing} • \textit{Uncertainty Estimation} • \textit{Theory of Mind} • \textit{Memory Recall}
    
    \item \textbf{Negotiation} A multi-player strategic trading game where players manage resources with different personal valuations. \\
    \textbf{Skills:} \textit{Persuasion} • \textit{Theory of Mind} • \textit{Strategic Planning}
    
    \item \textbf{Pit} A fast-paced card game simulating commodity trading. \\
    \textbf{Skills:} \textit{Uncertainty Estimation}
    
    \item \textbf{Poker} A card game where players bet on hand values while bluffing and reading opponents. \\
    \textbf{Skills:} \textit{Bluffing} • \textit{Uncertainty Estimation} • \textit{Theory of Mind} • \textit{Strategic Planning} • \textit{Persuasion}
    
    \item \textbf{Snake} A multi-player adaptation of the classic arcade game where players control a snake that grows by eating apples. \\
    \textbf{Skills:} \textit{Spatial Thinking} • \textit{Strategic Planning}
    
    \item \textbf{Surround} A classic multiplayer arcade game where players control a continuously moving line, aiming to trap opponents or avoid collisions. \\
    \textbf{Skills:} \textit{Spatial Thinking} • \textit{Strategic Planning}
    
    \item \textbf{Two Rooms and a Boom} A social deduction party game involving hidden roles and separate group interactions. \\
    \textbf{Skills:} \textit{Theory of Mind} • \textit{Persuasion} • \textit{Strategic Planning}
    
    \item \textbf{Diplomacy} A strategic board game focused on negotiation, alliances, and tactical movement. \\
    \textbf{Skills:} \textit{Persuasion} • \textit{Theory of Mind} • \textit{Strategic Planning} • \textit{Adaptability}
\end{itemize}

\section{Online Model Play}
\label{app:online_play}
The code setup might change over time. For the most up to date example, please check the \href{https://github.com/LeonGuertler/TextArena}{Github}.
\begin{lstlisting}%[language=Python]
import textarena as ta
 
model_name = "MODEL_NAME"
model_description = "MODEL_DESCRIPTION"
email = "EMAIL_ADDRESS"


# Initialize agent
agent = ta.agents.OpenRouterAgent(model_name="gpt-4o")  


env = ta.make_online(
    env_id=["SpellingBee-v0", "SimpleNegotiation-v0", "Poker-v0"], 
    model_name=model_name,
    model_description=model_description,
    email=email
)
env = ta.wrappers.LLMObservationWrapper(env=env)


env.reset(num_players=1)

done = False
while not done:
    player_id, observation = env.get_observation()
    action = agent(observation)
    done, info = env.step(action=action)


rewards = env.close()
\end{lstlisting}

\end{document}